%% file: main.tex
\documentclass[10pt,twocolumn,letterpaper]{article}

\usepackage[pagenumbers]{cvpr} 
\usepackage{marvosym}
\input{preamble}
\usepackage[table]{xcolor}
\usepackage{wrapfig}
\definecolor{lightgray}{RGB}{235,235,235}
\definecolor{myyellow}{RGB}{220, 220, 120}

\definecolor{cvprblue}{rgb}{0.21,0.49,0.74}
\usepackage[pagebackref,breaklinks,colorlinks,allcolors=cvprblue]{hyperref}

\def\paperID{***} 
\def\confName{CVPR}
\def\confYear{2026}

\title{The Prism Hypothesis: \\Harmonizing Semantic and Pixel Representations via Unified Autoencoding}

\author{Weichen Fan$^{1,2}$ \hspace{.2cm}  Haiwen Diao$^{1}$ \hspace{.2cm} Quan Wang$^{2}$ 
\hspace{.2cm} Dahua Lin$^{2}$ \hspace{.2cm}  Ziwei Liu$^{1,}$\textsuperscript{\Letter} \\
\textsuperscript{1}S-Lab, Nanyang Technological University \quad
\textsuperscript{2}SenseTime Research\\
{\tt\small weichen002@e.ntu.edu.sg, \ haiwen.diao@ntu.edu.sg,} \\
{\tt\small {\{wangquan,dhlin\}@sensetime.com, \ ziwei.liu@ntu.edu.sg}}\\[1ex]
\parbox{\textwidth}{
\centering
\begin{tabular}{ll}
\raisebox{-0.15em}{\includegraphics[height=1.05em]{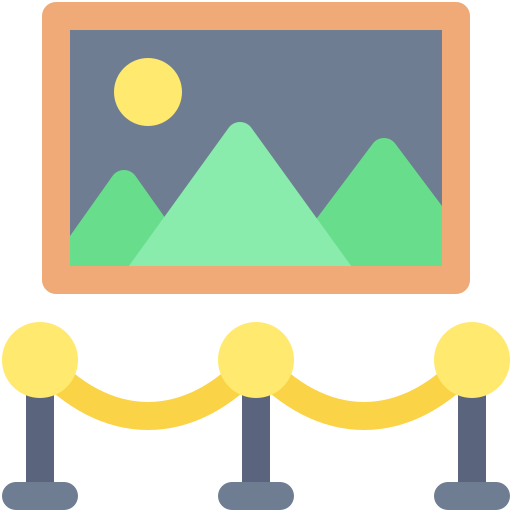}} \textbf{Github:} & \url{https://github.com/WeichenFan/UAE}. \\
\raisebox{-0.15em}{\includegraphics[height=1.05em]{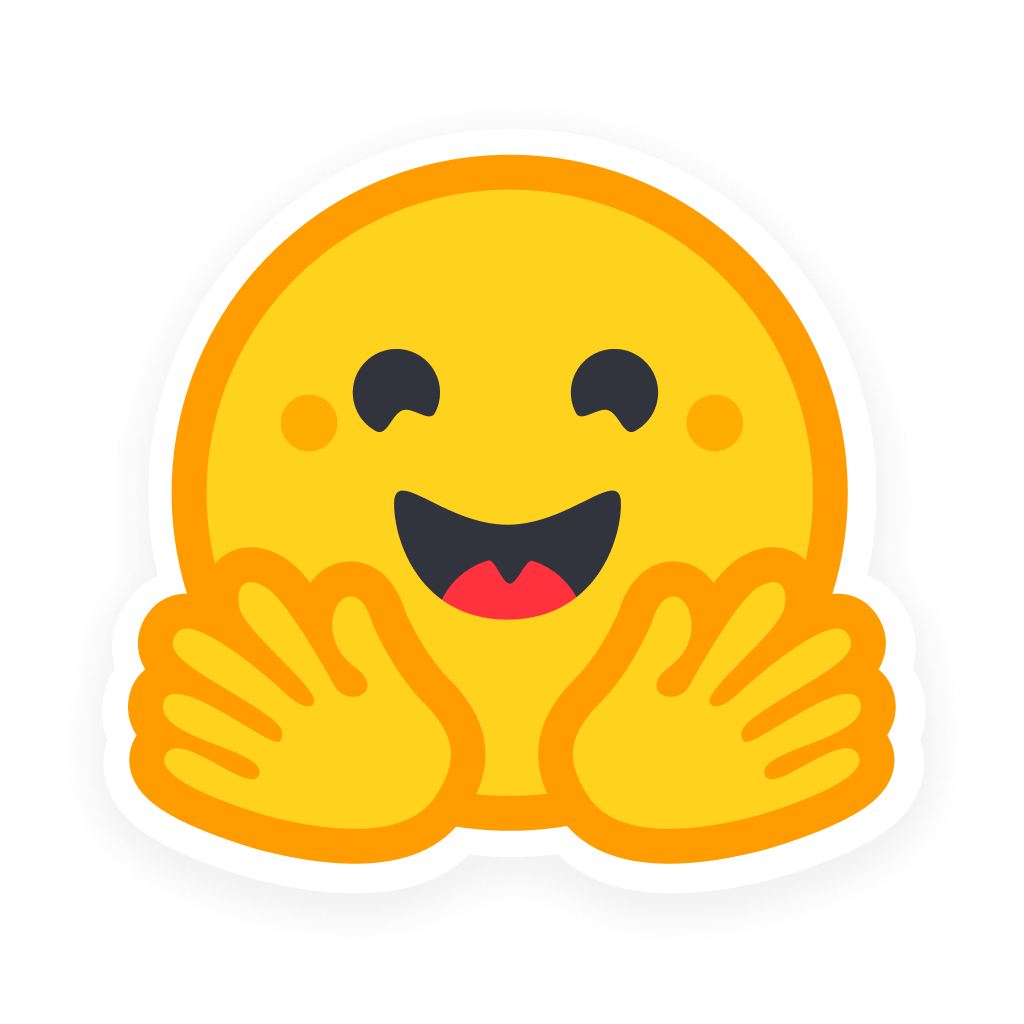}} \textbf{Hugging Face:} & \url{https://huggingface.co/weepiess2383/UAE}. \\
\end{tabular}
}
}

\begin{document}

\twocolumn[{%
\renewcommand\twocolumn[1][]{#1}%
\maketitle

\begin{center}
    \captionsetup{type=figure}
    \small
\includegraphics[width=0.94\linewidth,
    trim=5 85 20 62,
    clip]{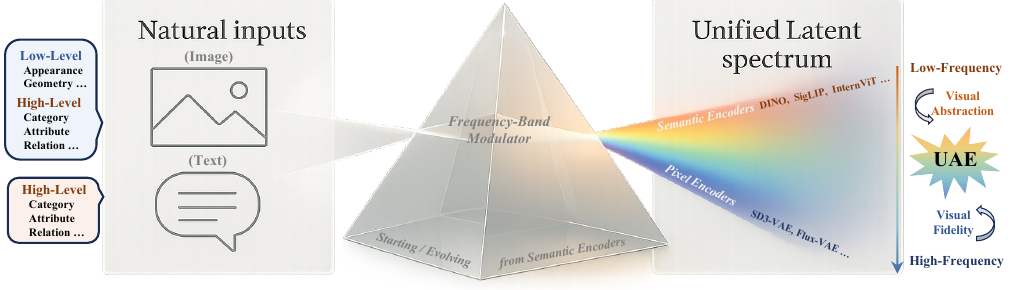}
    \caption{
    \textbf{The Prism Hypothesis.}
    Our conceptual “prism” decomposes various natural inputs into spectral components along frequency.
    Low frequency bands capture global semantics and abstract meaning, while high frequency bands encode local detail and fine visual texture.
    This motivates our Unified Autoencoding (UAE), which harmonizes semantic and pixel representations within a single latent space.
    }
    \label{fig:prism}
\end{center}}]

\input{sec/0_abstract}    
\input{sec/1_intro}
\input{sec/2_rel}
\input{sec/3_method}
\input{sec/4_exp}

\input{sec/ablation}
\input{sec/5_conclusion}
{
    \small
    \bibliographystyle{ieeenat_fullname}
    \bibliography{main}
}


\end{document}

%% file: sec/0_abstract.tex
\begin{abstract}
Deep representations across modalities are inherently intertwined. In this paper, we systematically analyze the spectral characteristics of various semantic and pixel encoders. Interestingly, our study uncovers a highly inspiring and rarely explored correspondence between an encoder’s feature spectrum and its functional role: semantic encoders primarily capture low-frequency components that encode abstract meaning, whereas pixel encoders additionally retain high-frequency information that conveys fine-grained detail. This heuristic finding offers a unifying perspective that ties encoder behavior to its underlying spectral structure.
We define it as \textbf{the Prism Hypothesis}, where each data modality can be viewed as a projection of the natural world onto a shared feature spectrum, just like the prism.
Building on this insight, we propose \textbf{Unified Autoencoding (UAE)}, a model that harmonizes semantic structure and pixel details via an innovative frequency-band modulator, enabling their seamless coexistence.
Extensive experiments demonstrate that UAE effectively unifies semantic abstraction and pixel-level fidelity within a single latent space, achieving state-of-the-art performance. Moreover, we show that UAE can be directly applied to pixel-space modeling, significantly improving both FID and IS over the vanilla JIT baseline.
\end{abstract}

%% file: sec/1_intro.tex
\vspace{-0.5cm}
\section{Introduction}
\label{sec:intro}

Trained on massive corpora, recent foundation models have profoundly reshaped perception and generation systems, generalizing well across diverse downstream tasks~\cite{dosovitskiy2021vit,caron2021dino,oquab2023dinov2,radford2021clip}.
Yet, early advances in perception and generation evolve along largely separate trajectories. 
Their objectives are typically distributed across distinct network structures, \eg, employing pretrained semantic encoders~\cite{caron2021dino,oquab2023dinov2,radford2021clip}, to capture high-level meaning, or pixel encoders~\cite{kingma2014vae,razavi2019vqvae2} to compress fine-grained visual detail.
While each module excels within its own domain, this fragmentation compels subsequent unification efforts~\cite{qu2025tokenflow,huang2025illume+,tang2025unilip} to depend simultaneously on semantic and pixel encoders, forcing networks to reconcile fundamentally heterogeneous representations.
The sharp mismatch lowers training efficiency and induces representational conflicts, with these incompatible features often interfering rather than complementing one another.

This fragmentation uncovers a deep-seated tension between abstraction and fidelity, which is a driving force in shaping subsequent foundation models.
To alleviate it, recent studies~\cite{zheng2025rae,yue2025uniflow,shi2025svg} attempt to transfer semantic encoders into visual generation domains alongside strong pixel decoders. 
This strategy substantially accelerates convergence and improves semantic correspondence, yet remains limited in recovering fine-grained visual details.
In parallel, another research direction seeks to endow pixel encoders with semantic awareness through 
text supervision~\cite{xu2022groupvit,yi2023simseg,wu2024imagetext}, 
semantic encoder distillation~\cite{chen2023zeroseg,huang2023g2sd}, 
and hierarchical feature integration~\cite{wu2024imagetext}.
While these efforts move toward unifying understanding and generation representations in a single module, they often achieve coexistence through trade-offs rather than genuine integration.

At the core of this development lies a fundamental question: How is information about the world represented such that multimodal inputs share a common semantic meaning while preserving their native granularity of detail? 

Impressively, we empirically observe that pre-trained semantic features, whether derived from text or vision, tend to reside at the coarse end of the decoupled feature spectrum, primarily capturing low-frequency structures such as categories, attributes, and relations.
In contrast, pre-trained pixel-level features extend toward the finer end of the spectrum, representing higher-frequency components that convey intricate appearance and geometric detail.
Strikingly, these complementary representations can be harmoniously integrated within a unified encoder, extremely aligning well with the spectral arrangement and fostering a progressive synergy from semantic perception to detailed reconstruction.
%
We posit this as \textbf{\textit{the Prism Hypothesis}}: as shown in Figure~\ref{fig:prism}, the real-world inputs project onto a continuous shared feature spectrum, and what we perceive as different modalities are distinct slices of this underlying continuum.

From this perspective, we introduce \textbf{Unified Autoencoding (UAE)}, a tokenizer that learns a shared latent space and harmonizes semantic structure and pixel-level detail.
Specifically, UAE features a frequency decomposition that factorizes real-world content into a fundamental semantic band and residual pixel bands with controllable fine granularity.
This design is not only grounded in comprehensive empirical evidence but also validated across diverse reconstruction and perception tasks.
With its compact yet semantically expressive representations, UAE outperforms concurrent RAE~\cite{zheng2025rae}, SVG~\cite{shi2025svg}, and UniFlow~\cite{yue2025uniflow} across rFID, PSNR, gFID and Accuracy metrics on ImageNet, demonstrating that its learned latent space is both semantically representative and pixel-faithful.

Our work makes the following contributions.
\begin{enumerate}
\item We introduce the \textbf{Prism Hypothesis}, a spectral perspective that explains multimodal data through a shared fundamental band and modality-specific bands, supported by extensive empirical observations.

\item We propose \textbf{Unified Autoencoding (UAE)}, a simple yet effective tokenizer via frequency transformation that integrates seamlessly with diffusion transformers and enables flexible generative modeling in the latent space.

\item UAE is effective in \textbf{both pixel and latent spaces}. In the latent space, UAE achieves competitive performance in both generative modeling and pixel-level reconstruction. In the pixel space, integrating UAE into JIT~\cite{li2025back} leads to faster convergence and improved FID and IS.
\end{enumerate}

%% file: sec/2_rel.tex

\section{Related Work}
\label{sec:related_work}

\paragraph{Unified Tokenizers and Unified Representations.}

Unifying the representations between pixel and semantic embeddings has become a central objective for existing foundation models. Joint embedding approaches align images and text in a shared representation and enable strong zero-shot transfer \cite{radford2021clip,jia2021align}, and have been extended to many modalities, \eg, audio, depth, thermal, and inertial signals \cite{girdhar2023imagebind}. In parallel, modality-agnostic backbones aim to build a unified architecture that can process diverse input modalities and generate task-specific outputs through learned queries or a shared token representation \cite{jaegle2021perceiverio,lu2022unifiedio,lu2024unifiedio2,wang2023beit3,baevski2022data2vec}.  

On the tokenizer side, discrete codebook methods have demonstrated that the design and granularity of visual tokens play a crucial role in determining how effectively a single sequence model can adapt to vision tasks~\cite{esser2021taming,yu2024magvitv2}. Recent work goes further and seeks tokenizers that support both understanding and generation at the same time. OmniTokenizer~\cite{wang2024omnitokenizer} learns a joint image video tokenizer with a spatial-temporal decoupled transformer and reports strong reconstruction and synthesis across both domains. Very recent studies deepen this trend. UniFlow~\cite{yue2025uniflow} proposes a unified pixel flow tokenizer that adapts a pretrained encoder through layer-wise self-distillation and employs a lightweight patch-wise flow decoder, explicitly targeting the long-standing tension between semantic abstraction and pixel-faithful reconstruction. Two concurrent works remove the traditional variational bottleneck entirely and build unified representation latents for diffusion transformers. Diffusion Transformers with Representation Autoencoders (RAE)~\cite{zheng2025rae} replace the usual reconstruction-only encoder with a pretrained representation encoder, \eg, DINO or SigLIP, and a trained decoder, arguing that semantically rich latents accelerate convergence and improve generative fidelity. SVG~\cite{shi2025svg} trains a diffusion model on DINO features with a small residual branch for details, reporting faster training, few-step sampling, and improved quality.  

Our UAE aligns with this shift. It serves as a unified tokenizer that decouples continuous latent features explicitly corresponding to the underlying spectral structure by factorizing real-world contents into a low-frequency base and residual high-frequency bands. 
It anchors semantics in the core representation while relegating fine-grained details to residuals for progressive reconstruction.

\paragraph{Frequency and Multi-Resolution Modeling.}
Classical image synthesis adopted pyramids and wavelets to separate structure by scale, enabling coarse-to-fine generation and targeted refinement of detail. A typical example is the Laplacian pyramid of adversarial networks~\cite{denton2015lapgan}, which trains a generator per level and synthesizes images by successively adding higher-frequency residuals. Subsequent analyses of neural networks from a spectral perspective showed that standard architectures prioritize low frequencies and learn higher frequencies later, a phenomenon known as spectral bias~\cite{rahaman2019spectral}. Two lines of work respond to this bias. The first uses input or architecture design to improve access to high frequencies, \eg, Fourier feature mappings and periodic activations that help multi-layer perceptrons represent fine detail~\cite{tancik2020fourier,sitzmann2020siren}. The second introduces frequency-aware objectives and signal processing choices, such as focal frequency loss to emphasize hard frequencies and alias-free synthesis to avoid spurious high-frequency artifacts~\cite{jiang2021ffl,karras2021aliasfree}.

\input{figs/energy}

Modern generative models retain this multi-resolution view. Cascaded diffusion trains models at increasing resolutions, allowing each stage to learn the appropriate frequency band and its own error distribution~\cite{ho2022cascaded}. Variants construct explicit feature pyramids or hierarchical patch schedules, enabling efficient diffusion on large images and video while preserving high-frequency detail~\cite{fan2023frido,skorokhodov2024hpdm}. Recent latent diffusion designs introduce cross-magnification spaces or zoomable pyramids that share information across scales and enable large-image reconstruction without retraining~\cite{yellapragada2025zoomldm}. 
In autoregressive models, VAR~\cite{tian2024var} casts generation as predicting the next scale or resolution, showing strong ImageNet results and clean scaling trends. This explicit progression from global layout to fine detail emerges as a viable alternative to diffusion.
Building on this idea, Next Visual Granularity (NVG) generation~\cite{wang2025next} produces sequences at a fixed resolution but with progressively finer token granularity, surpassing prior VAR baselines while maintaining structured control over detail. Similarly, NFIG~\cite{huang2025nfig} performs discrete next-frequency prediction, demonstrating strong generative performance on the ImageNet benchmark.
In pixel space, DCTdiff~\cite{ning2024dctdiff} explicitly trun image into DCT space and perform generative modeling.
Our approach aligns with these trends but focuses on unified representation rather than the generator’s training schedule.

%% file: figs/energy.tex
\begin{figure}[t]
\centering
\includegraphics[width=0.95\linewidth]{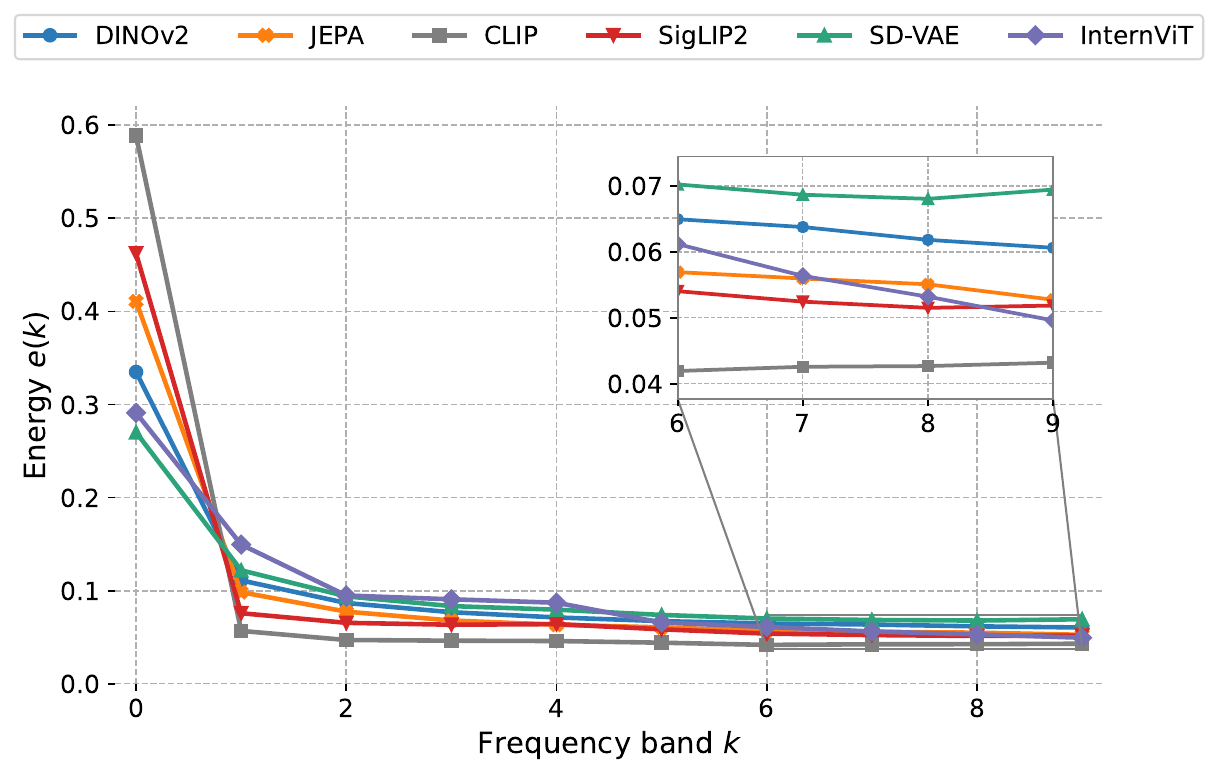}
\caption{
\textbf{Frequency energy distribution.}
Normalized energy $e(k)$ across frequency bands for diverse tokenizers.
DINOv2 and CLIP focus on low-frequency (semantic) content, while SD-VAE retains more high-frequency energy, capturing finer details.
}
\label{fig:energy}
\end{figure}

%% file: sec/3_method.tex
\section{Methodology}
\label{sec:method}

\input{figs/prsim_exp}
\input{figs/overview}

\subsection{Preliminary Findings}
\label{sec:prelim}

\noindent \textbf{The Prism Hypothesis.}
Natural inputs are regarded as projections of a common real-world signal onto a shared frequency spectrum. Semantic encoders emphasize a compact low-band that carries categories, attributes, and relations, while pixel encoders observe the same fundamental base together with higher bands that encode edges, textures, and fine appearance. The hypothesis indicates that cross-modal alignment depends primarily on the shared low band.

\noindent \textbf{Formalization.}
Here, \(\mathcal{F}\) and \(\mathcal{F}^{\dagger}\) denote two-dimensional discrete Fourier transform and its inverse. 
For an image \(I\in[0,1]^{3\times H\times W}\) and a smooth radial mask \(\mathbf{M}^{\mathrm{LP}}_{\rho}\) that passes frequencies within normalized radius \(\rho\in(0,1]\),
\begin{equation}
I^{\mathrm{LP}}_{\rho}\;=\;\mathcal{F}^{\dagger}\!\bigl(\mathbf{M}^{\mathrm{LP}}_{\rho}\odot\mathcal{F}(I)\bigr),
\end{equation}
\begin{equation}
I^{\mathrm{HP}}_{\rho}\;=\;\mathcal{F}^{\dagger}\!\bigl(\mathbf{M}^{\mathrm{HP}}_{\rho}\odot\mathcal{F}(I)\bigr),
\end{equation}
where \(\mathbf{M}^{\mathrm{HP}}_{\rho}\) is complementary to \(\mathbf{M}^{\mathrm{LP}}_{\rho}\), and both masks use cosine transitions to limit ringing artifacts.
All filtering is performed in linear space prior to any model-specific normalization.
With a frozen vision-language encoder \(E\), we compute cosine similarities between text embeddings and image embeddings from \(I\), \(I^{\mathrm{LP}}_{\rho}\), and \(I^{\mathrm{HP}}_{\rho}\).
If semantic alignment is carried by the shared base, then the retrieval score \(\mathcal{R}\) satisfies $\mathcal{R}_{\mathrm{LP}}(\rho)\ \text{is nondecreasing in}\ \rho,$
$\mathcal{R}_{\mathrm{HP}}(\rho)\ \text{is nonincreasing in}\ \rho$, and \(\mathcal{R}_{\mathrm{HP}}(\rho)\) approaches chance once the shared base is removed.

\noindent\textbf{Empirical Verification.}
We empirically examine the Prism Hypothesis using two complementary analyses: (i) token-space frequency energy decomposition and (ii) robustness of text--image alignment under controlled spectral filtering.

\textbf{Exp1 (Token-spectrum energy).}
For each encoder, we extract patch tokens from the input images, reshape them into a 2D token grid, and apply a channel-wise 2D DCT to obtain token-frequency coefficients. 
We then reorder the coefficients via zig-zag traversal (progressing from low to high frequencies), partition them into coarse bands, and compute the normalized band energy $e(k)$ as the average of squared magnitudes across channels and images.
As shown in Fig.~\ref{fig:energy} (left), the resulting spectra are dominated by the lowest band, revealing a strong low-frequency bias across models (which is consistent with the natural images). 
Importantly, different encoders allocate different \emph{relative} mass to higher bands: 
Semantic representation encoders concentrate more energy in low-frequency components, while the pixel-centric SD-VAE preserves relatively more energy in the mid- and high-frequency bands (inset), consistent with its stronger retention of fine-grained appearance details.

\textbf{Exp2 (Low-pass retrieval).}
To directly investigate the frequency localization of cross-modal semantics, we evaluate text–image retrieval under progressive removal of high-frequency image components. Concretely, we apply a radial low-pass mask in the frequency domain, with a cutoff specified as a fraction of the Nyquist frequency, reconstruct the filtered images, and measure Recall@1 using frozen vision–language models (CLIP and SigLIP2) with fixed text prompts
(\textit{an image of ...}).
Fig.~\ref{fig:retrieval_cutoff} (right) shows that Recall@1 increases rapidly when only a small low-frequency portion is retained (e.g., $\sim$0.23 at 0.05 Nyquist vs.\ $\sim$0.58 at 0.10), and saturates near full-image performance by moderate cutoffs (around 0.3--0.4 Nyquist).
This behavior indicates that a significant fraction of retrieval-relevant semantics is encoded in coarse, low-frequency image structure, whereas higher-frequency components are largely unrelated to semantic information.

Taken together, Exp1 and Exp2 support the central intuition of the Prism Hypothesis: cross-modal semantic alignment is primarily associated with a shared low-frequency base, whereas higher-frequency bands increasingly encode modality-specific, fine-grained visual detail.

\input{figs/dit_model}
\input{figs/recon}

\subsection{Unified AutoEncoder}
Using DINOv2 as an example, we initialize UAE from the pretrained encoder and discard the register tokens, retaining only the patch tokens with channel dimension $C$.
Input images are center-cropped and resized to the encoder’s native input resolution. The resulting patch-token sequence is then reshaped into a 2D latent grid
$\mathbf{z}\in\mathbb{R}^{B\times C\times H\times W}$,
which enables explicit frequency-domain processing.

\noindent\textbf{Frequency transform and tokenization.}
Given $\mathbf{z}$, we apply a channel-wise 2D discrete cosine transform (DCT) $f(\cdot)$ to obtain a frequency-domain representation
\begin{equation}
\mathbf{h} = f(\mathbf{z}), \qquad \mathbf{h}\in\mathbb{R}^{B\times C\times H\times W}.
\end{equation}
We then linearize the 2D DCT grid into a 1D token sequence via a frequency-preserving reordering scheme.

\vspace{0.5ex}
\noindent\textbf{Token Reorder.}
In the 2D DCT grid, each coefficient $C(u,v)$ (not to be confused with the channel dimension $C$) corresponds to a spatial frequency indexed by $(u,v)$. A natural measure of frequency magnitude is the radial norm
\begin{equation}
r(u,v) = \sqrt{u^2 + v^2}.
\end{equation}
As $(u,v)$ moves away from the top-left DC component $(0,0)$, $r(u,v)$ increases in discrete levels, yielding progressively higher spatial frequencies. This allows us to interpret the DCT lattice as a set of concentric ``radial shells'' centered at $(0,0)$.

To convert the 2D frequency grid into a 1D token sequence while approximately preserving low-to-high frequency progression, we adopt zig-zag reordering. Zig-zag traversal roughly sorts coefficients by increasing $u+v$, which serves as a practical proxy for radial growth (since $u^2+v^2$ typically increases with $u+v$ for small-to-moderate indices). This produces an ordered sequence
\begin{equation}
\{ C_k \}_{k=0}^{N-1}, \qquad N = H \times W,
\end{equation}
where the frequency magnitude increases approximately monotonically with $k$.

\vspace{0.5ex}
\noindent\textbf{Semantic Regularization.}
To preserve the semantic priors of the pretrained teacher while expanding the representational bandwidth, we apply a \emph{semantic-wise} alignment loss only on the lowest-frequency bands. Let $\mathbf{f}_s$ denote features from the frozen pretrained semantic encoder (teacher), and $\mathbf{f}_u$ denote features from the trainable unified encoder (student). After frequency decomposition, we obtain band-wise representations $\{\mathbf{f}^k_s\}_{k=0}^{K-1}$ and $\{\mathbf{f}^k_u\}_{k=0}^{K-1}$.
We enforce alignment only for the first $K_{\text{base}}$ low-frequency bands, which primarily encode global structure and category-level semantics, while leaving higher frequency less constrained:
\begin{equation}
\mathcal{L}_{\text{sem}} =
\frac{
\sum_{k=0}^{K_{\text{base}}-1}
\|\mathbf{f}^k_u - \mathbf{f}^k_s\|_2^2}{K_{\text{base}}}.
\end{equation}
This restricted supervision transfers the teacher’s semantic organization to the low-frequency subspace, while leaving higher-frequency bands unconstrained so they can specialize in modality-specific, pixel-level detail. Empirically, this selective regularization stabilizes training and mitigates collapse toward purely pixel-driven features.

\vspace{0.5ex}
\noindent\textbf{Frequency Masked Prediction.}
During training, we randomly mask a subset of high-frequency tokens to encourage robust inference of missing fine details. Let $\mathcal{L}$ and $\mathcal{H}$ denote the index sets of low- and high-frequency components, respectively. We construct a masked frequency representation $\tilde{\mathbf{F}}$ as
\begin{equation}
\tilde{\mathbf{F}}_{i} =
\begin{cases}
\mathbf{F}_{i}, & i \in \mathcal{L}, \\
0, & i \in \mathcal{H}_{\text{mask}},
\end{cases}
\end{equation}
where $\mathcal{H}_{\text{mask}} \subseteq \mathcal{H}$ is sampled independently at each iteration. That is, selected high-frequency coefficients are set to zero, while low-frequency components remain intact.

The masked frequency tokens are mapped back to the spatial domain via inverse DCT and fed to the decoder. 
The decoder is trained to reconstruct the full high-fidelity image, including the masked high-frequency details, conditioned only on the preserved low-frequency structure and the remaining visible coefficients. This objective encourages the model to infer fine-grained textures and high-frequency signals from global semantic cues, improving robustness and generalization.

\input{tabs/reconstruction}
\subsection{Generative Modeling}
As illustrated in Figure~\ref{fig:overview}, UAE latents admit two complementary parameterizations for generative modeling: \emph{spatial-domain} features and \emph{frequency-domain} coefficients. 
In \textbf{spatial-wise} modeling, we first apply the inverse DCT to map the frequency tokens back to a spatial latent grid, and then train a standard latent diffusion transformer on these spatial features, following prior representation-based tokenizers~\cite{zheng2025rae}.
In \textbf{frequency-wise} modeling, we operate directly on the reordered DCT coefficients. Since our decoder can reconstruct high-quality images even when conditioned only on low-frequency tokens, we can adopt a coarse-to-fine generation strategy: the diffusion transformer is trained to model low-frequency components first and is then progressively extended to incorporate higher-frequency bands, yielding increasingly fine-grained details.

%% file: figs/prsim_exp.tex
\begin{figure}[t]
\centering
\includegraphics[width=0.90\linewidth]{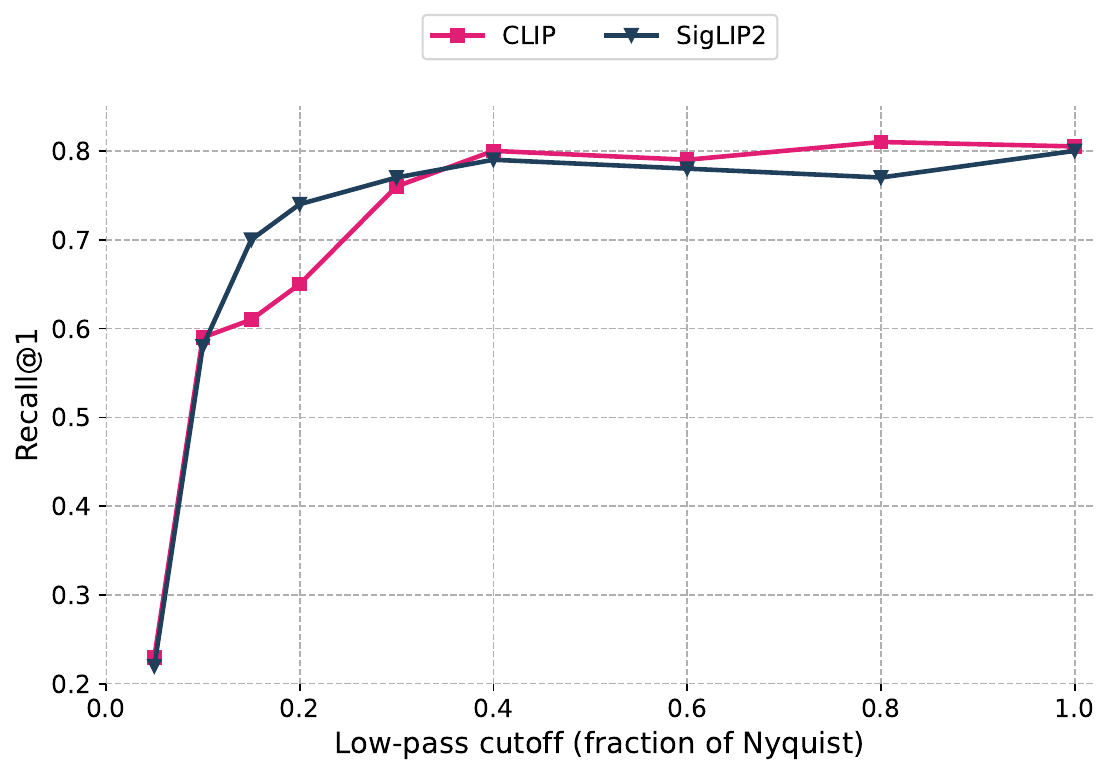}
\vspace{-2.0mm}
    \caption{
    \textbf{Retrieval results via frequency filtering.}
    Text–Image retrieval remains stable under low-pass filtering but degrades sharply under high-pass filtering,
    confirming that semantic alignment primarily resides in low-frequency components. 
    }
\label{fig:retrieval_cutoff}
\end{figure}

%% file: figs/overview.tex
\begin{figure*}[t]
\centering
\includegraphics[width=0.99\linewidth]{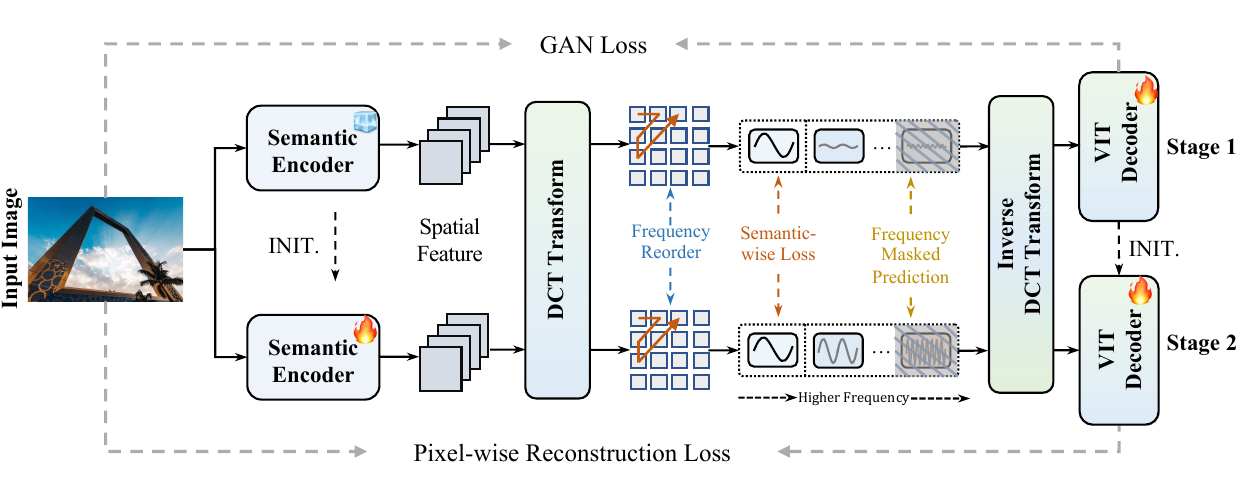}
\vspace{-18pt}
    \caption{
    \textbf{Overall architecture of our proposed Unified Autoencoding (UAE).} 
    The input image is separately encoded by both a pretrained \emph{Semantic Encoder} (e.g., DINOv2) and the trainable \emph{Unified Encoder}. 
    The unified encoder is initialized from the semantic encoder and optimized under two complementary objectives:
    a \textbf{semantic-wise loss} that aligns low-frequency components decomposed from the semantic encoder’s representations, 
    and a \textbf{pixel-wise reconstruction loss} that enforces visual fidelity via the \emph{Pixel Decoder} by adaptively dilating the high-frequency components. 
    The decoder employs spectral transform blocks to refine residual-frequency content and produce the reconstructed image. 
    This joint optimization harmonizes semantic structure and pixel detail within a single latent space.}
\label{fig:overview}
\vspace{-2mm}
\end{figure*}

%% file: figs/dit_model.tex
\begin{figure*}[t]
\centering
\includegraphics[width=0.99\linewidth]{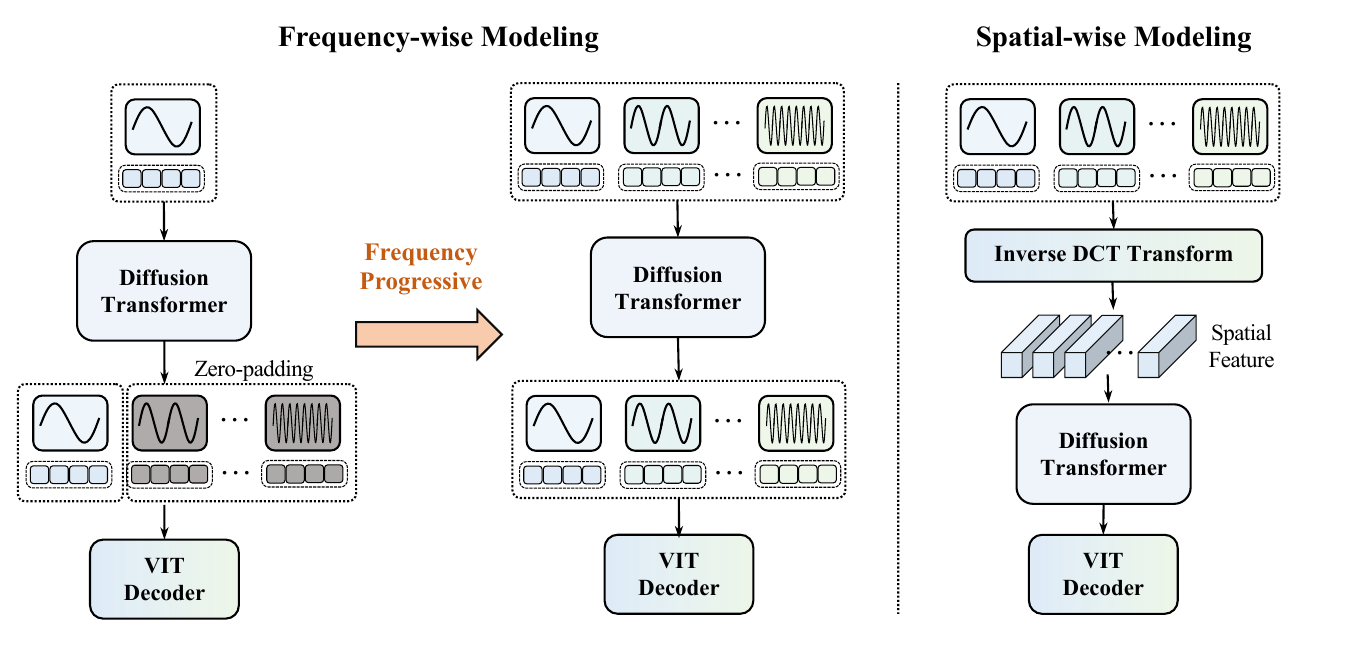}
\caption{
\textbf{Frequency-wise vs. spatial-wise modeling.}
\emph{Left:} The diffusion transformer operates directly in the DCT domain and progressively models frequency tokens from low to high bands before decoding. 
\emph{Right:} Frequency tokens are first transformed back to spatial features via inverse DCT, and diffusion is performed in the spatial domain prior to decoding.
}
\label{fig:overview}
\end{figure*}

%% file: figs/recon.tex
\begin{figure*}[t]
\centering
\includegraphics[width=0.99\linewidth]{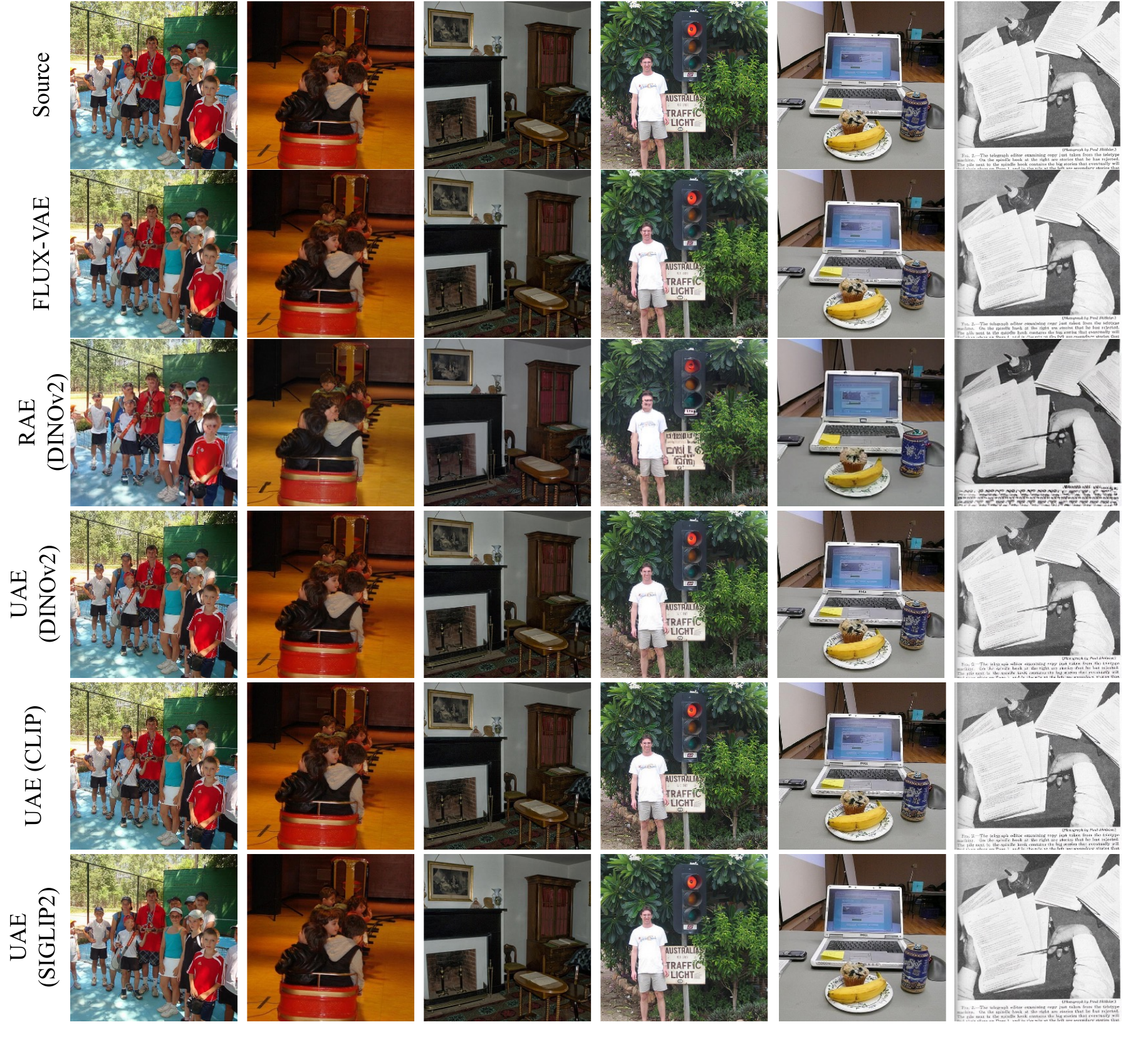}
\caption{
\textbf{Qualitative comparison of reconstruction fidelity across autoencoding paradigms.}
We visualize reconstructed samples from representative methods, including Flux-VAE~\cite{labs2025flux1kontextflowmatching}, RAE~\cite{zheng2025rae}, and our proposed UAE.
Each row corresponds to reconstructions from a fixed source set spanning text, human, object, and artistic domains.
UAE produces the most consistent and semantically faithful reconstructions, preserving both high-frequency details (e.g., texture and edge sharpness) and global structure (e.g., layout and color harmony), while reducing the blurring and semantic drift observed in RAE. 
}
\label{fig:recon}
\end{figure*}

%% file: tabs/reconstruction.tex
\begin{table*}[t]
\centering
\caption{
\textbf{Reconstruction performance on ImageNet-1K and MS-COCO 2017.}
UAE consistently achieves strong PSNR/SSIM and competitive rFID compared with both continuous autoencoders and unified tokenizers.
We additionally report linear probing accuracy to evaluate representation quality.
}
\resizebox{0.98\textwidth}{!}{
\begin{tabular}{lcccccccccc}
\toprule
\textbf{Method} & \textbf{Type} & \textbf{Ratio} & \textbf{Linear prob (Acc)$\uparrow$} &
\multicolumn{3}{c}{\textbf{ImageNet-1K}} &
\multicolumn{3}{c}{\textbf{MS-COCO 2017}} \\
\cmidrule(lr){5-7} \cmidrule(lr){8-10}
 & & &  & 
 \textbf{PSNR$\uparrow$} & \textbf{SSIM$\uparrow$} & \textbf{rFID$\downarrow$} & 
 \textbf{PSNR$\uparrow$} & \textbf{SSIM$\uparrow$} & \textbf{rFID$\downarrow$} \\
\midrule

SD-VAE-1.x~\cite{rombach2021highresolution} & Continuous & 8 & -- 
& 23.54 & 0.68 & 1.22 
& 23.21 & 0.69 & 5.94 \\

SD-VAE~\cite{rombach2021highresolution} & Continuous & 8 & -- 
& 25.68 & 0.72 & 0.75 
& 25.43 & 0.73 & 0.76 \\

SD-VAE-2.x~\cite{rombach2021highresolution} & Continuous & 8 & -- 
& 23.54 & 0.68 & 1.22 
& 26.62 & 0.77 & 4.26 \\

SD-VAE-XL~\cite{podell2023sdxl} & Continuous & 8 & -- 
& 27.37 & 0.78 & 0.67 
& 27.08 & 0.80 & 3.93 \\

SD-VAE-3~\cite{lopez2025sd3} & Continuous & 8 & -- 
& 31.29 & 0.87 & 0.20 
& 31.18 & 0.89 & 1.67 \\

FLUX-VAE~\cite{labs2025flux1kontextflowmatching} & Continuous & 8 & -- 
& \textbf{32.74} & \textbf{0.92} & \textbf{0.18} 
& \textbf{32.32} & \textbf{0.93} & \textbf{1.35} \\

VA-VAE & Continuous & 16 & -- 
& 27.96 & 0.79 & 0.28 
& 27.50 & 0.81 & 2.71 \\

\midrule

SVG (DINOv3)~\cite{shi2025svg} & Continuous & 16 & --  
& 24.25 & 0.67 & 0.78 & - & - & - \\ 

RAE (DINOv2-B)~\cite{zheng2025rae} & Continuous & 14 & -- 
& 18.05 & 0.5 & 2.04 
& 18.36 & 0.47 & 6.01 \\ 

UniFlow (DINOv2-L)~\cite{yue2025uniflow} & Continuous & 14 & -- 
& 32.32 & 0.91 & 0.17 
& 30.66 & 0.94 & 2.81 \\ 

\rowcolor{lightgray}
\textbf{UAE (Siglip2)} & Continuous & 16 & \textbf{80.1 (81.2)}
& \textbf{31.00} & \textbf{0.91} & \textbf{0.43} 
& \textbf{30.20} & \textbf{0.89} & \textbf{2.91} \\ 

\rowcolor{lightgray}
\textbf{UAE (DINOv2-B)} & Continuous & 14 & \textbf{83.0 (83.0)}
& \textbf{32.17} & \textbf{0.92} & \textbf{0.35} 
& \textbf{31.19} & \textbf{0.91} & \textbf{2.01} \\ 

\rowcolor{lightgray}
\textbf{UAE (CLIP-L)} & Continuous & 14 & \textbf{77.2 (79.4)} 
& \textbf{36.58} & \textbf{0.96} & \textbf{0.04} 
& \textbf{36.25} & \textbf{0.97} & \textbf{0.41} \\ 

\bottomrule
\end{tabular}
}
\label{tab:token_comp}
\end{table*}

%% file: sec/4_exp.tex
\section{Experiments}
\label{sec:method}

\subsection{Implementation Details}

\noindent \textbf{UAE Training.}
We train UAE with DINOv2-B, DINOv2-L, SigLIP2, and InternViT at $256 \times 256$ resolution.
Training is performed in two stages.
In \textbf{Stage~1}, we freeze the pretrained semantic encoder and train the decoder using a pixel-wise reconstruction loss and an adversarial loss, following the original RAE setting~\cite{zheng2025rae}.
We use AdamW with an initial learning rate of $2\times10^{-4}$, linearly decayed to $2\times10^{-5}$, and train for 16 epochs.
The discriminator is optimized with AdamW using the same learning-rate schedule; discriminator training starts at epoch~6, and the GAN loss term on the generator is enabled from epoch~8. In \textbf{Stage~2}, we unfreeze the encoder and fine-tune the full model using the semantic-wise loss together with the reconstruction loss.
We additionally apply frequency-masked prediction by randomly masking high-frequency components and training the model to reconstruct the full image under the GAN objective.
The masking ratio is set to 75\% for DINOv2 and 50\% for the other encoders.
We set $K_{\text{base}}$ to 25\% of the total tokens for DINOv2 and 50\% for the remaining models.Unless otherwise specified, Stage~2 uses the same optimizer and learning-rate schedule as Stage~1, with high-frequency masking enabled from the beginning of Stage~2.

\noindent \textbf{Generative Modeling.}
We train SiT on two UAE latent parameterizations: spatial latents ($768 \times 16 \times 16$) and frequency latents ($768 \times 256$).
In both cases, training is performed progressively over frequency tokens: we start from 64 tokens for the first 80 epochs, then linearly increase the token count to 256, updating the token budget every 20 epochs.
All models are trained for a total of 1400 epochs to ensure stable convergence.
We adopt the AdamW optimizer with an initial learning rate of $2\times10^{-4}$, which is linearly decayed to $2\times10^{-5}$ over the course of training.

\subsection{Visual Reconstruction}
\label{sec:visual_recon}

\noindent \textbf{Quantitative Evaluation.}
In~\Cref{tab:token_comp}, We quantify reconstruction quality at \(256 \times 256\) on ImageNet-1K and MS-COCO 2017. The compared baselines include widely-used generative tokenizers and variational decoders. For a fair comparison, we report the evaluation results for our UAE and the corresponding baseline that follows the same configurations of DINOv2-base. Here, we report PSNR and SSIM for fidelity, and rFID for perceptual quality.

Notably, our UAE delivers state-of-the-art reconstruction quality among unified tokenizers. On ImageNet-1K, UAE improves over the RAE baseline from \(18.05\) to \(31.00\) in PSNR and from \(0.50\) to \(0.92\) in SSIM, while reducing FID from \(2.04\) to \(0.35\). On MS-COCO, UAE raises PSNR from \(18.36\) to \(31.19\) and SSIM from \(0.47\) to \(0.91\), with FID decreasing from \(6.01\) to \(2.01\). These notable gains validate that moving masking out of the decoder and factorizing latents into a low-frequency base and residual bands preserve fine detail while maintaining semantic structure.

Beyond unified tokenizers, UAE is competitive with the best generative-only tokenizers. On ImageNet-1K, UAE(CLIP-L) attains an FID of \(0.04\), surpassing SD3 VAE and approaching Flux VAE, while maintaining high PSNR and SSIM. A similar pattern holds on MS-COCO, where UAE(CLIP-L) surpasses the strongest baselines in perceptual quality.

\noindent \textbf{Qualitative Evaluation.}
Figure~\ref{fig:recon} compares source images with the resulting reconstructions from FLUX-VAE, RAE, and UAE. Note that our UAE can well preserve straight edges, fine textures, and small text, \eg, street signs and printed documents. The improvement is consistent across natural photos and illustrations. Moreover, we demonstrate that UAE is effective across different semantic encoders, including DINOv2~\cite{oquab2023dinov2}, CLIP~\cite{radford2021clip}, and SigLip2~\cite{tschannen2025siglip}.

\subsection{Generative Modeling}
\input{tabs/visual_gen}
To further assess the effectiveness of the proposed UAE, we conduct class-conditional image generation experiments on ImageNet-1K at a resolution of $256\times256$. 

We conduct all generative experiments in the multi-band latent space. All experimental settings follow those used in RAE~\cite{zheng2025rae} to ensure a fair comparison.
As shown in Table~\ref{tab:visual_gen}, our UAE attains a gFID of \textbf{1.52} and an IS of \textbf{234.5}, achieving performance on par with existing state-of-the-art generative models.
This suggests that the unified frequency-based representation enables the generative model to progressively capture both global structure and fine-grained details in a coherent manner, keeping highly generative while preserving strong semantic quality.
Overall, the UAE latent space, constructed through explicit low- and high-frequency decomposition, provides an effective and diffusion-friendly foundation for large-scale visual generation.

%% file: tabs/visual_gen.tex
\begin{table}[t]
\centering
\caption{
\textbf{Class-conditional generation performance on ImageNet (256x256).}
We compare our proposed UAE with recent diffusion and autoregressive models using standard metrics. Note that the UAE performs generation in a causal manner, progressing from low- to high-frequency bands in the latent space.
}
\resizebox{0.45\textwidth}{!}{
\begin{tabular}{lcccc}
\toprule
\textbf{\cellcolor{white}Methods} & \textbf{gFID$\downarrow$} & \textbf{IS$\uparrow$} & \textbf{Prec$\uparrow$} &
\textbf{Rec$\uparrow$}\\
\midrule
DiT~\cite{peebles2023dit}  & 9.62 & 121.5 & 0.67 & 0.67 \\
SiT~\cite{ma2024sit}     & 8.61 & 131.7 & 0.68 & 0.67 \\
SVG~\cite{shi2025svg} & 3.36 & 181.2 & - & - \\
UniFlow~\cite{yue2025uniflow} & 2.45 & 228.0 & - & - \\
RAE~\cite{zheng2025rae} & \textbf{1.51} & \textbf{242.9}& \underline{0.79} & \textbf{0.63} \\
\midrule
\rowcolor{lightgray}
UAE & \underline{1.52} & \underline{234.5} & \textbf{0.81} & \underline{0.62} \\

\bottomrule
\end{tabular}}
\label{tab:visual_gen}
\end{table}

%% file: sec/ablation.tex
\section{Discussion}
\label{sec:ablation}
\input{figs/fid_ablation}
\subsection{The Impact of Generative Modeling}
We further study the effect of generative modeling using UAE representations. As shown in Fig.~\ref{fig:fid_converge}, UAE significantly accelerates diffusion model convergence compared with SiT and RAE. Both spatial and frequency variants consistently achieve lower FID throughout training and reach high-quality generation within substantially fewer epochs. In particular, UAE attains strong performance as early as 40 epochs and continues to improve steadily, while baseline methods converge much more slowly. The zoom-in view highlights the final performance gap at 80 epochs, where UAE achieves the best FID. These results suggest that the frequency-aware representation of UAE provides a more generation-friendly latent space, enabling more efficient generative modeling.
\input{tabs/efficiency}
\input{tabs/semantic_mask}
\subsection{Inference Efficiency}
We compare the computational efficiency of different tokenizer and latent modeling configurations in Table~\ref{tab:compute_latency}. 
The proposed UAE tokenizer introduces negligible overhead in stage-1 reconstruction, exhibiting nearly identical GFLOPs and latency to the RAE baseline. In contrast, the efficiency gain becomes significant in stage-2 generative modeling. When the SiT backbone operates on the compact UAE 64 low-freq tokens, the computational cost is reduced from 313.84 GFLOPs to 78.80 GFLOPs per image, resulting in a $\sim$4$\times$ reduction, while inference latency decreases from 8.82 ms to 4.05 ms per image.

\subsection{The Impact of Semantic Loss}
We ablate the strength of semantic regularization by varying $K_{\text{base}}$ as shown in the Table~\ref{tab:ablation_sem_mask} (a), the fraction of low-frequency tokens constrained by the semantic alignment loss using DINOv2 as the teacher. 
When semantic regularization is disabled ($K_{\text{base}}=0$), the model achieves strong reconstruction but exhibits weak semantic alignment, as reconstruction gradients favor local appearance cues. 
Increasing $K_{\text{base}}$ improves semantic performance by anchoring the low-frequency latent space to the pretrained semantic geometry, but excessive constraints reduce reconstruction fidelity. 
We find that aligning only 25\% of frequency tokens already preserves most semantic capability while maintaining high reconstruction quality, supporting the design that semantics mainly reside in the low-frequency base while higher frequencies capture fine details.

\input{figs/energy_comp}
\subsection{The Impact of Mask Predictions}
We further study the effect of frequency-masked prediction during UAE training by varying the masking ratio applied to high-frequency tokens. This strategy serves as a regularizer by forcing the decoder to reconstruct missing high-frequency details from the preserved low-frequency structure and remaining visible components. As shown in Table~\ref{tab:ablation_sem_mask} (b), aggresive masking yields the best generative performance: masking 85\% of high-frequency tokens achieves the lowest gFID@20 epochs (9.7). When the masking ratio is too small, the regularization effect is limited and the model tends to rely on deterministic high-frequency cues. In contrast, larger masking ratios remove excessive high-frequency information and significantly improve the converge speed. These results suggest that strong masking encourages stronger structural priors in the latent space and leads to more effective generative modeling.

\subsection{The Impact of Unified Training}

As shown in Fig.~\ref{fig:energy_comp}, we analyze the frequency energy distribution of vanilla encoders and their unified-trained counterparts under the same setting described in Sec.~\ref{sec:prelim}.
Unified training induces a consistent redistribution of energy across frequency bands.
Specifically, the UAE models exhibit reduced dominance in low-frequency components, accompanied by a noticeable increase in mid- and high-frequency energy.
This shift suggests that unified training enhances the representation of fine-grained details while largely preserving global semantic structure.
The zoom-in view further confirms that high-frequency components are better retained or even amplified after unification.

\input{tabs/fid_pixel}

\subsection{UAE in Pixel Space}
In latent space, we observe that a pretrained semantic encoder concentrates most of the useful information in low-frequency components, while fine-grained image details are primarily encoded in high-frequency tokens. As shown in Table~\ref{tab:semantic_emerge}(a), we conduct generative modeling while retaining different numbers of low-frequency tokens. The results demonstrate that generative performance strongly favors low-frequency tokens (i.e., semantic components), which is consistent with recent findings~\cite{baade2026latent}.

In Table~\ref{tab:semantic_emerge}(b), we further train an encoder–decoder model from scratch while preserving only $N$ low-frequency tokens, and perform linear probing on the learned representations. The results indicate that semantic information naturally emerges during the compression and reconstruction process.

These observations suggest that reconstruction and generation are intrinsically connected through the extent to which low-frequency tokens are preserved. From this perspective, pixel-space modeling and latent-space modeling are fundamentally equivalent, differing only in compression ratio and the specific compression architecture (e.g., SD-VAE with $8\times$ compression is analogous to JIT with patch size 8). This insight allows us to directly extend UAE to pixel-space modeling.

Therefore, We further evaluate the effectiveness of unified training in the pixel-space generative setting by applying UAE to the JIT framework.
As shown in Table~\ref{tab:fid_pixel}, we compare the original JIT model with its unified-trained counterpart under identical configurations. 
UAE consistently outperforms the vanilla JIT across all metrics and training stages.
UAE achieves a notable improvement, consistently achieve lower fid score and high IS score. These results indicate that unified training extends beyond latent space and remains effective when applied directly in pixel space.

\input{tabs/semantic_emerge}

%% file: figs/fid_ablation.tex
\begin{figure}[t]
\centering
\includegraphics[width=\columnwidth]{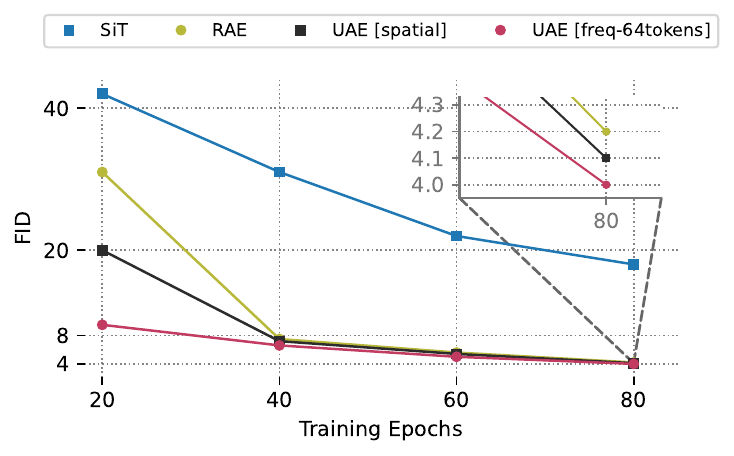}
\caption{UAE (spatial and frequency) converges significantly faster and achieves lower FID than SiT and RAE. The zoom-in highlights the final performance gap at 80 epochs.}
\label{fig:fid_converge}
\end{figure}

%% file: tabs/efficiency.tex
\begin{table}[t]
\centering
\caption{
\textbf{Efficiency comparison across different representations.}
UAE maintains comparable computational cost to RAE in both spatial and frequency variants, while enabling significant efficiency gains when integrated with SiT.
In particular, \textit{SiT (UAE-freq-64tokens)} achieves a 4$\times$ reduction in GFLOPs and over 2$\times$ lower latency compared to the RAE-based baseline.
}
\resizebox{0.45\textwidth}{!}{
\begin{tabular}{lcc}
\toprule
\textbf{Model} & \textbf{GFLOPs} & \textbf{Latency} \\
\midrule
RAE & 266.24 & 8.40 \\
\rowcolor{lightgray}
UAE-spatial & 266.24 & 8.40 \\
\rowcolor{lightgray}
UAE-freq & 266.27 & 8.76 \\
\midrule
SiT (RAE) & 313.84 & 8.82 \\
\rowcolor{lightgray}
SiT (UAE-freq-64tokens) & \textbf{78.80} & \textbf{4.05} \\
\bottomrule
\end{tabular}}
\label{tab:compute_latency}
\end{table}

%% file: tabs/semantic_mask.tex
\begin{table}
\centering
\caption{\textbf{Ablations on semantic loss and mask prediction.}}

\begin{minipage}[t]{0.49\linewidth}
\centering
\textbf{(a)}\\
\vspace{1mm}
\begin{tabular*}{\linewidth}{@{\extracolsep{\fill}}lcc}
\toprule
$K_{base}$ & PSNR$\uparrow$ & ACC$\uparrow$ \\
\midrule
0\%  & 34.3 & 60.1 \\
25\% & 32.2 & 83.0 \\
50\% & 28.1 & 83.0 \\
75\% & 20.1 & 83.0 \\
\bottomrule
\end{tabular*}
\end{minipage}
\hfill
\begin{minipage}[t]{0.49\linewidth}
\centering
\textbf{(b)}\\
\vspace{1mm}
\begin{tabular*}{\linewidth}{@{\extracolsep{\fill}}lc}
\toprule
Ratio & gFID@20 epoch$\downarrow$ \\
\midrule
85\% & 9.7 \\
15\% & 12.2 \\
50\% & 27.2 \\
25\% & 33.1 \\
\bottomrule
\end{tabular*}
\end{minipage}

\label{tab:ablation_sem_mask}
\end{table}

%% file: figs/energy_comp.tex
\begin{figure}[t]
\centering
\includegraphics[width=0.95\linewidth]{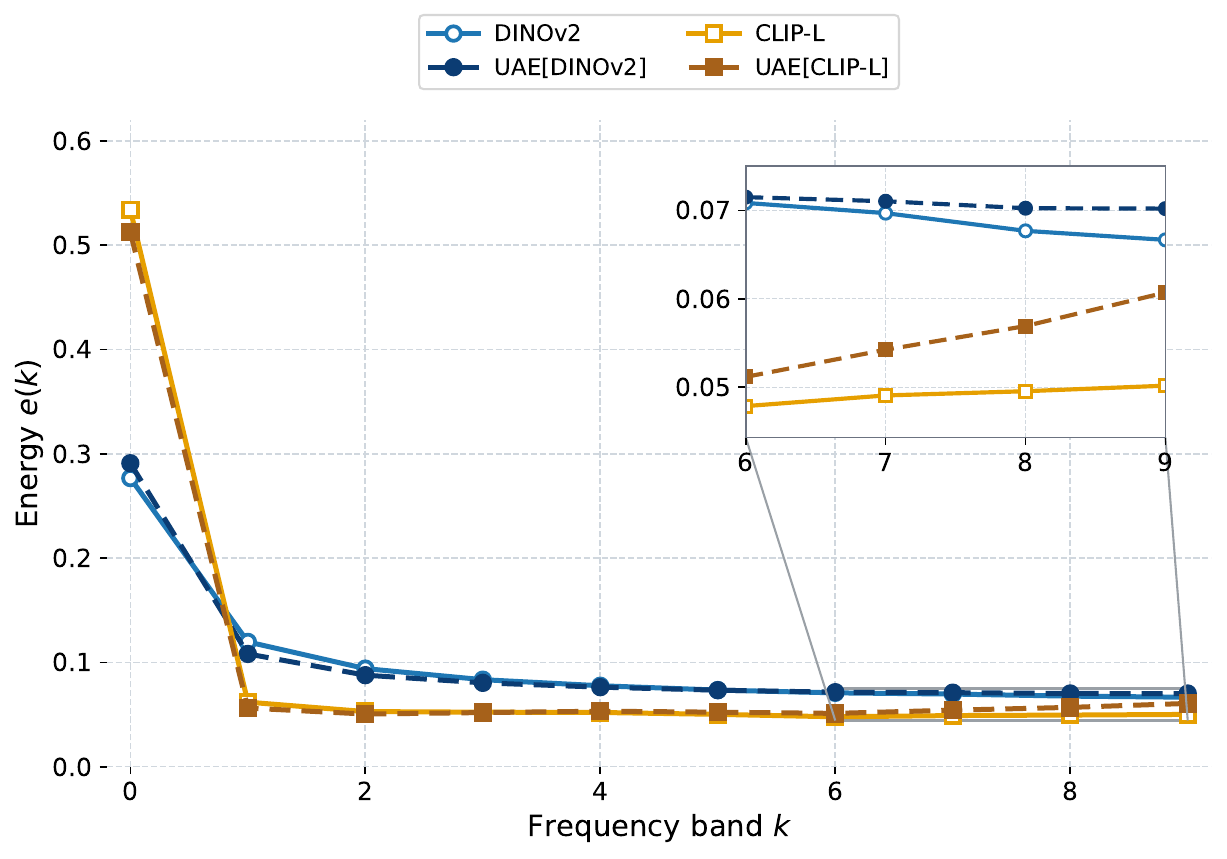}
\caption{
\textbf{Energy distribution across frequency bands before and after unified training.}
Unified training consistently shifts energy from low-frequency components to mid- and high-frequency bands.
The zoom-in (right) highlights that UAE preserves or enhances high-frequency energy, indicating improved fine-grained detail representation, while only slightly reducing low-frequency dominance associated with global semantics.
}
\label{fig:energy_comp}
\end{figure}

%% file: tabs/fid_pixel.tex
\begin{table}[t]
\centering
\caption{
\textbf{Comparison of generative performance in pixel space.}
We evaluate JIT and its unified-trained variant UAE (JIT) under the same architecture, model size, and patch size.
UAE consistently improves both FID ($\downarrow$) and IS ($\uparrow$) across training stages.
}
\resizebox{0.45\textwidth}{!}{
\begin{tabular}{lccccc}
\toprule
\textbf{Model} & \textbf{Epoch} & \textbf{Size} & \textbf{Patch Size} & \textbf{FID$\downarrow$} & \textbf{IS$\uparrow$} \\
\midrule
JIT~\cite{li2025back} & 40 & 700M & 32 & 31.81 & 60.13 \\
JIT~\cite{li2025back} & 80 & 700M & 32 & 16.06 & 123.30 \\
\midrule
\rowcolor{lightgray}
\textbf{UAE (JIT)} & 40 & 700M & 32 & \textbf{25.80} & \textbf{72.82} \\
\rowcolor{lightgray}
\textbf{UAE (JIT)} & 80 & 700M & 32 & \textbf{14.55} & \textbf{132.57} \\
\bottomrule
\end{tabular}}
\label{tab:fid_pixel}
\end{table}

%% file: tabs/semantic_emerge.tex
\begin{table}
\centering
\caption{(a). Latent space generative modeling converge speed with different low-freq tokens. (b). Linear-probing accuracy with models trained from scratch in different low-freq tokens.}

\begin{minipage}[t]{0.49\linewidth}
\centering
\textbf{(a)}\\
\vspace{1mm}
\begin{tabular*}{\linewidth}{@{\extracolsep{\fill}}lcc}
\toprule
Tokens & Epochs & FID$\downarrow$ \\
\midrule
64  & 20 & 9.7 \\
81 & 20 & 12.2 \\
100 & 20 & 20.6 \\
121 & 20 & 40.5 \\
\bottomrule
\end{tabular*}
\end{minipage}
\hfill
\begin{minipage}[t]{0.49\linewidth}
\centering
\textbf{(b)}\\
\vspace{1mm}
\begin{tabular*}{\linewidth}{@{\extracolsep{\fill}}lc}
\toprule
Tokens & Accuracy$\uparrow$ \\
\midrule
64 & 60.1 \\
81 & 33.2 \\
100 & 28.1 \\
121 & 20.7 \\
\bottomrule
\end{tabular*}
\end{minipage}

\label{tab:semantic_emerge}
\end{table}

%% file: sec/5_conclusion.tex
\section{Conclusion}
\label{sec:conclusion}

We propose the Prism Hypothesis, which views diverse natural inputs as projections of a shared spectrum composed of a compact low-frequency semantic component and residual higher-frequency detail.
Preliminary experiments about frequency-band distributions of various visual and textual encoders strongly enlighten this connection between feature spectra and representational function.
Hence, we launch Unified AutoEncoding (UAE), which harmonizes semantic and pixel information within a single latent space via a hierarchical frequency-band modulator.
Extensive experiments confirm that it delivers greater generative capability over existing unified tokenizers, \eg, RAE, SVG, and UniFlow, meanwhile preserving reconstruction quality on par with top-tier models like Flux-VAE. We further validate that the idea of UAE not only benefit in the latent space but also remains effective in pixel space.
We position the UAE as a promising and practical route towards unified tokenizers for understanding and generation.
\textcolor{white}{.}